%% file: main.tex
\documentclass[letterpaper]{article} 
\usepackage[preprint]{aaai2027}
\usepackage[hyphens]{url}  
\usepackage{graphicx} 
\usepackage{natbib}  
\usepackage{caption} 
\usepackage{booktabs}
\usepackage{array}
\usepackage{tabularx}
\usepackage{amsmath}
\usepackage{amssymb}

\newcommand{\vf}{\mathrm{VF}}
\newcommand{\fcr}{\mathrm{FCR}}
\title{SimVerity: When Does Simulated Agent Success Survive Physical Deployment?}
\author{
    Zhonghao Zhan\textsuperscript{\rm 1},
    Krinos Li\textsuperscript{\rm 1},
    Yefan Zhang\textsuperscript{\rm 2},
    Hamed Haddadi\textsuperscript{\rm 1}
}
\affiliations{
    \textsuperscript{\rm 1}Imperial College London\\
    \textsuperscript{\rm 2}Independent Researcher
}

\begin{document}
\maketitle

\input{sections/abstract}

\input{sections/introduction}

\input{sections/verdict_transfer}

\input{sections/experimental_design}

\input{sections/evidence_and_calibration}

\input{sections/held_out_evaluation}

\input{sections/related_work}

\input{sections/limitations_ethics_reproducibility}

\input{sections/conclusion}

\bibliography{simverity_aaai2027}

\end{document}

%% file: sections/abstract.tex
\begin{abstract}
Simulated evaluation is widely used to benchmark AI agents, yet how much evidence a simulated pass provides about physical deployment has not been systematically quantified. We present SimVerity, a verdict-transfer assurance framework: it replays matched scenarios on target smart home deployments and cross-validates agent execution against independently qualified physical witnesses. Our evaluation highlights that deployment success is a real-world process, not a static property in simulation: completion, reported state, observable effect, and settled outcome diverged within the same execution. Although an advanced simulator cleared all 240 light trials, a camera caught 42 sub-second failures invisible to settled-state checks. False clearance was predictable: a risk profile learned from measured trials and locked before evaluation predicted failures on a path it never physically measured, beating a property-blind baseline in all eleven held-out sessions across two cohorts. Agent auditability was also measurable: switching one agent loop's model--client/serving configuration raised its scenario-matching share from 52--88\% to 100\%. Finally, a second qualified simulator added no independent cross-check: it never disagreed on any overlapping case, and only physical measurement exposed their shared blind spots. SimVerity turns verdict transfer into an explicit decision: clear, abstain, or escalate before deployment.
\end{abstract}

%% file: sections/introduction.tex
\section{Introduction}
\label{sec:introduction}

As autonomous AI agents transition into physical domains, they now command real homes.  Google Home's agent executes
blinds and lights as one spoken command, and lets
camera scene understanding trigger automations across the home
\citep{google2026gemini}.  Before such an agent ships, its green check
comes from simulation: smart home benchmarks provide executable
devices and state-based checks \citep{seo2026simuhome,
rivkin2024sage,li2025homebench,li2026smhbench,gu2026homeflow}, tool
agents receive sandbox verdicts \citep{yao2024taubench}, and world
models score rollouts \citep{quevedo2025evaluating,li2025worldeval}.
A fundamental question remains unaddressed: does a simulator's pass
justify physical deployment?  When it does not, the false clearance,
an approval whose property then fails, is an oversight failure, not
benchmark noise \citep{kapoor2024agents,xue2025illusion,ruan2024toolemu}.

The root cause of false clearance is an abstraction mismatch:
simulation sees success as a static property, while physical
deployment unfolds as a process.  The stakes are ordinary: in a basic agentic smart-home routine,
the camera sees the car leave and the
house shuts itself down, lights off, the space heater's smart plug
off.  Where stakes are obvious, vendors hard-code defenses: smart
locks verify their own bolt, relock on 30-second timers, and are
polled every 30 seconds by Home Assistant
\citep{schlage2026autolock,utec2026autolock,ha2026schlage}. However, a heater
on a plug has only its reported state and whatever happens to be
watching the room.  Because false clearance cannot be rehearsed on
real burners, we measure the same anatomy where failure is harmless:
an agent issues \texttt{turn\_off} and checks after 100 ms, the
declared \emph{read boundary}.  The simulator passed all five held-out
\texttt{off} trials and software state usually agreed, yet a qualified
camera still saw the light on in 5/5; at the settled boundary, all 30
trials passed, including those 5.  Completion, reported state,
observable effect, and settled effect had been collapsed into one
word: success.  In SimuHome itself, a 3-second evaluation delay moves
one agent's light-transition success from 44\% to 62\%
\citep[App.~K]{seo2026simuhome}.  Success depends on when and where it
is observed.

\begin{figure}[t]
  \centering
  \includegraphics[width=1.0\columnwidth]{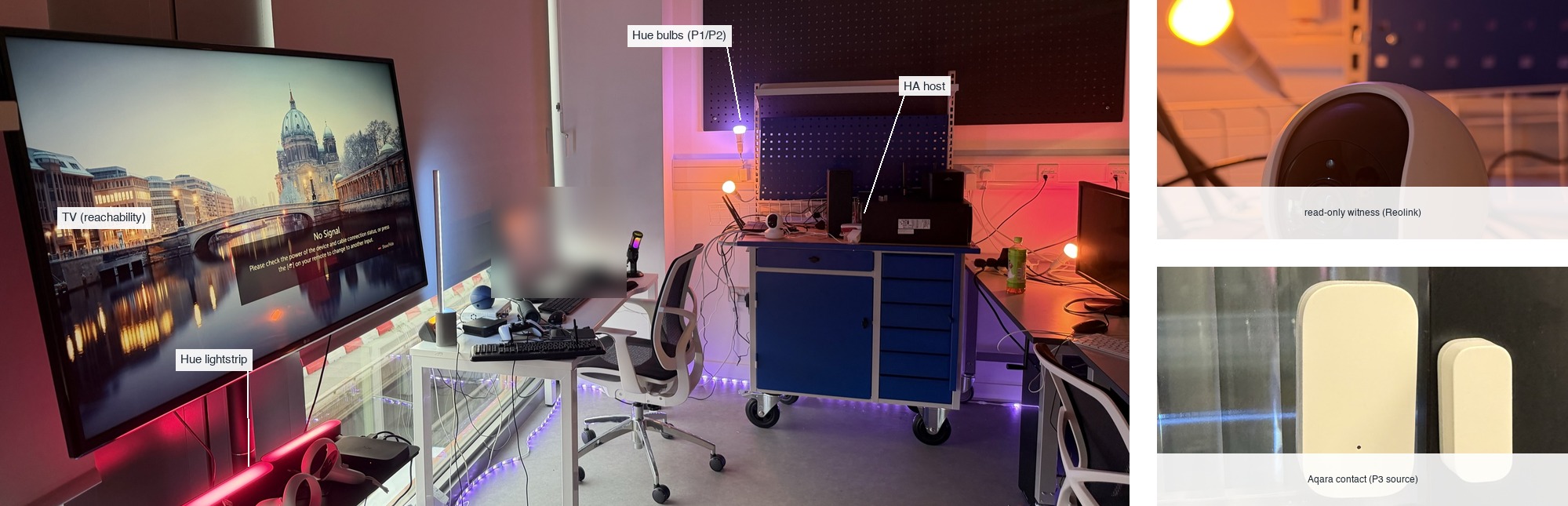}
  \caption{Instrumented Smart Home deployment.}
  \label{fig:testbed}
\end{figure}

Existing verification paradigms fall short of this property--verdict
discrepancy.  Simulation methodology insists a model is valid only for
a stated purpose \citep{sargent2013verification}; conformance theory
for cyber-physical systems proves which properties carry from model to
system, but under dynamics assumptions that semantic, stochastic agent
traces do not satisfy \citep{abbas2014formal,roehm2019conformance};
robotics and driving studies ask whether simulations predict reality, not whether one specific pass survives
\citep{kadian2020predictivity,li2024simpler,fremont2020formal}.
Closest in spirit, an admissibility ladder names
``verdict-transfer-validated'' as its highest level but leaves it
unimplemented for lack of real-world anchors
\citep{oefinger2026validate}.  None supplies a reusable audit at
property-verdict granularity.

\begin{figure*}[t]
  \centering
  \includegraphics[width=0.7\textwidth]{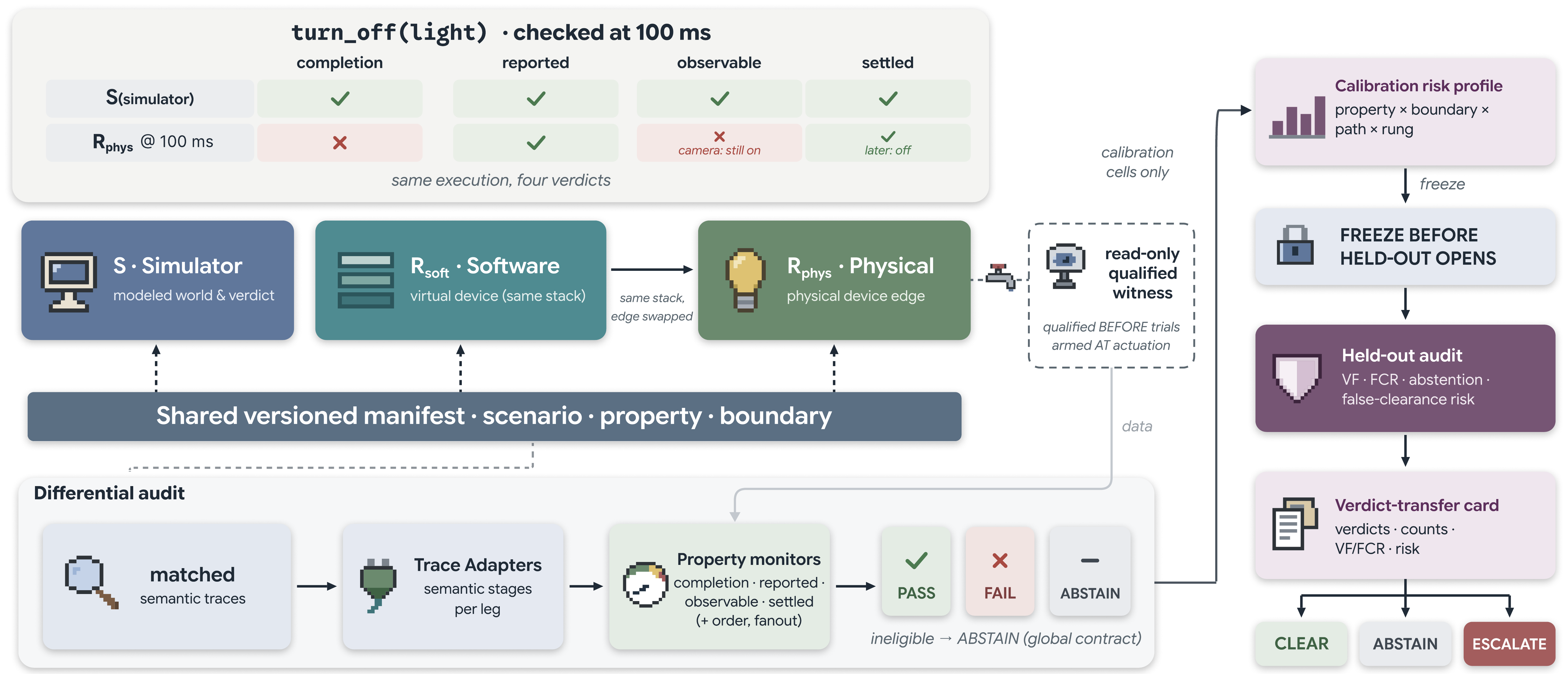}
  \caption{SimVerity audits a declared property across the simulator,
$R_{\mathrm{soft}}$ (real stack, virtual edges), and
$R_{\mathrm{phys}}$ (physical edges, prequalified read-only witness
armed at actuation); a shared manifest aligns matched traces, and
calibration-only cells freeze the risk profile before the held-out
audit returns clear, abstain, or escalate.}
  \label{fig:ladder}
\end{figure*}

To bridge this gap, we introduce SimVerity, an engine-agnostic
\emph{verdict-transfer assurance layer} outside the agent harness that
quantifies whether a simulator's verdict survives on the target
deployment.  It reruns the declared scenarios on the deployment and
grades each matched trace pair against the same declared property.  It
takes physical verdicts only from independently qualified witnesses (a
camera must first prove it can tell the two outcomes apart);
unqualified or missing evidence abstains rather than counting as
agreement.  From calibration runs alone it learns a frozen risk
profile predicting which simulator passes will fail in deployment.
The output is a card: risk, coverage, and abstain-or-escalate
dispositions indexed by property $\times$ read boundary $\times$ path
$\times$ deployment rung, the ladder Figure~\ref{fig:ladder} draws
from simulator to physical hardware.  The layer never chooses actions
or updates policy, and its witnesses are armed at actuation.  We do
not claim to have invented agent assurance or evaluator auditing; we
identify verdict transfer as a missing assurance contract: paired,
property-conditioned Verdict Fidelity and False Clearance Risk under
hash-bound provenance.

We validate SimVerity on physical smart-home testbeds: three findings
and a cross-check follow.  First, deployed success is a process in the
real world, not a property in simulation.  Second, false clearance is
predictable, and the simulator's own verdict is its worst predictor:
a risk profile frozen before evaluation beat a
property-blind baseline in all eleven held-out sessions across two
cohorts, under natural ambient variation and curtain-controlled
regimes alike; a deployment-local lookup stays stronger under
stationary control.  Third, auditability is measurable and
configuration-conditioned: an unmodified production harness, Hermes
\citep{nousresearch2026hermes}, kept every simulated trace matched to
the declared scenarios; two custom agent loops, one ReAct-style, one
planner-style, fell to 52--88\%; and a registered
model--client/serving intervention restored the same ReAct loop to
100\%.  Meanwhile, a second simulator changed coverage without ever
disagreeing, leaving deployment anchors as the only independent check.

%% file: sections/verdict_transfer.tex
\section{Methods}
\label{sec:method}

\subsection{Problem Formulation}
\label{sec:problem}

Let $x\sim\mathcal{D}$ be a legal scenario drawn from a declared
evaluation distribution and $\phi$ a declared trace property.  A
\emph{source} rung $Q$ issues the clearance under audit, and a
\emph{target} rung $T$ is where its survival is measured; after
semantic alignment,
$V_Q^\phi(x),V_T^\phi(x)\in\{\mathrm{pass},\mathrm{fail},
\mathrm{abstain}\}$.  For binary eligible pairs, Verdict Fidelity is
\begin{equation}
  \vf_\phi = \Pr_{x\sim\mathcal{D}}
  [V_Q^\phi(x)=V_T^\phi(x)],
\end{equation}
and the deployment-critical error is False Clearance Risk,
\begin{equation}
  \fcr_\phi = \Pr[V_T^\phi(x)=\mathrm{fail}\mid
  V_Q^\phi(x)=\mathrm{pass}].
\end{equation}
Abstain is a first-class outcome: it is assigned when a witness fails qualification, a trace stage is missing, or a capability is
unsupported, and abstaining pairs are excluded from the binary
estimands but tallied in the audit accounting, so missing evidence
cannot silently become agreement.  This paper instantiates three
source--target pairings: direct simulator-to-physical calibration
($S\to R_{\mathrm{phys}}$), software-rung-to-physical held-out
prediction ($R_{\mathrm{soft}}\to R_{\mathrm{phys}}$), and
retrospective replay of a second simulator against frozen physical
anchors; the verdict-source column of
Table~\ref{tab:transfer-ledger} names $Q$ for every evidence block.
The interface is engine-agnostic: a learned world model, the judge
class the admissibility ladder targets \citep{oefinger2026validate},
is eligible as $Q$ whenever its adapter exposes the manifest's
actions, observations, and verdict-bearing property; present
evidence uses public executable simulators, so that transfer remains
an interface scope claim, not an empirical result.  These estimands
are conditional, not universal: a reported value answers one
question, under the exact coordinates that produced it, from scenario
distribution and read boundary to rungs and witness; change any
coordinate and the value must be re-earned.
This is operational validity at verdict granularity: the measured
complement of a priori conformance guarantees, which are unavailable
for semantic, stochastic agent traces.

\subsection{System Overview}

Figure~\ref{fig:ladder} shows the audit design.  A versioned
manifest fixes the scenario and every coordinate of its execution,
from read boundary and rungs to agent mode and split assignment.  As scenarios
execute on the source and target rungs, adapters collect traces
without modifying either engine, alignment maps them onto semantic
stages, witnesses qualify before testifying, and property monitors
grade each matched pair; every audited cell exports a
verdict-transfer card carrying $\vf_\phi$, $\fcr_\phi$, intervals, and a disposition: clear, abstain, or escalate. The contract is frozen before it is tested: every monitor,
threshold, and risk profile is locked under SHA-256 hashes before any
held-out ledger opens;
ledgers are immutable, and the evidence chain terminates at explicit, falsifiable measurement assumptions.

\subsection{Pipeline Components}

\paragraph{Trace adapters and semantic alignment.}
Pairing is a bijection on trial identity: extra or missing trials must raise errors.  Alignment maps each raw
trace onto six semantic stages (request, source or ingestion report,
agent read, dispatch, software feedback, observable effect);
unavailable stages remain unavailable, and an ingestion time is never
relabeled as the physical source time of a sensor event or effect.

\paragraph{Property monitors.}
A monitor is a total map from an aligned trace pair to
$\{\mathrm{pass},\mathrm{fail},\mathrm{abstain}\}$.  The frozen set
evaluates reported completion, reported read-after-write, observable
effect after write, settled postcondition, effect order, and fanout
completion; each declares its observation channel (reported state or
qualified witness) and abstains, rather than passing, on a missing
completion or boundary read, an invalid or missing witness, tied or
missing effect times, or any unwitnessed fanout target.
Sensor-to-effect scenarios reuse the read, observable, and settled
monitors with a declared sensor cause; a cross-device cell asks
whether an actuation leaves another device's reported channel quiet.

\paragraph{Qualified witnesses and session validity.}
Instruments qualify before they testify.  Let $\tilde L_{\mathrm{on}},
\tilde L_{\mathrm{off}}$ be median frame brightness over five settled
frames per reference state; an optical witness qualifies only if
\mbox{$|\tilde L_{\mathrm{on}}-\tilde L_{\mathrm{off}}| \ge
\max(10,\;20\,\widehat{\mathrm{MAD}}_{\mathrm{pooled}})$}, with
midpoint and direction then frozen.  Multi-target sessions further
require, under randomized actuation, every absolute cross-response
below half the target's own.  A session is valid only with trace
completeness ${\ge}95\%$ and settled-control agreement; invalid
initial states and missing witnesses abstain, and raw frames are
reduced to scalar region statistics at capture and discarded.  These
thresholds qualify the instrument and never tune a boundary after
outcomes. For example, the first pilot session during the experiment was invalidated by camera exposure drift and remains an audit record.

\paragraph{Calibration-only risk profiles.}
For each conditioning class $c$ (property $\times$ stage or direction
$\times$ path), the profile estimates deployment-failure risk from
source-cleared calibration rows as the Jeffreys-smoothed rate
$\hat r_c=(k_c+\tfrac12)/(n_c+1)$, with $n_c$ cleared pairs and $k_c$
their deployment failures; held-out cells without support back off
along a frozen conditioning ladder.  Brier score (squared error of
forecast risk) is primary because the output is a probability and
calibration matters \citep{brier1950verification}; log loss, AUROC,
coverage, and reliability plots are secondary.  The design is
intentionally simple: the question is whether declared structure
carries across cells (property--boundary--path--rung combinations),
not whether a learner can fit a testbed.  The initial
cohort's registered gate required beating a strong per-device latency
lookup and a path-only profile on the held-out path--rung pairing;
confirmatory cohorts registered path-only as primary criterion.

\subsection{Audit Dispositions and Decision Logic}

Every estimate ships with its full accounting: every clearance,
block, and abstention is counted, session and path totals stay
visible, and exact Clopper--Pearson intervals accompany each rate
\citep{clopper1934use}.  The
accounting blocks two cheats: a judge that rarely clears and an audit
that quietly discards hard-to-witness trials both show near-perfect
agreement, and the carried counts expose them as collapsed clearance
or abstention.  The intervals are trial-level summaries, and the
visible session and path counts delimit external inference: repeated
trials on one deployment are not independent deployments. An assurance layer invites the question of who assures the assurer;
the answer is that every element of the contract can fail visibly,
and absent qualified evidence the audit abstains.  Each audited cell exports a versioned
verdict-transfer card that binds the run's identities and hashes to verdicts and counts, and closes with one of three dispositions:
clear when evidence sustains the source's clearance,
abstain when evidence fails qualification, escalate when cleared
passes are observed to fail.

%% file: sections/experimental_design.tex
\section{Experimental Design}
\label{sec:design}

\subsection{Physical and Simulated Testbed}

The audit runs on a multi-rung testbed: one primary simulator and one
live deployment at two rungs.  The simulator $S$ is SimuHome,
unmodified and driven only through an external adapter; the
deployment runs Home Assistant on a live home, with the software rung
$R_{\mathrm{soft}}$ routing commands through virtual device edges in
the same stack and the physical rung $R_{\mathrm{phys}}$ terminating
at real commodity devices (smart lights, battery contact sensors)
under an independently qualified camera witness
(Figure~\ref{fig:testbed}; see supplement for details).  Four frozen paths carry the evidence: single-light effect (P1),
multi-light order/fanout (P2), contact-sensor-to-lamp (P3), and
cross-device observation (P5c, an append-only amendment within the
read-after-write property).  A path is reported as
ineligible when its payload, source provenance, or witness cannot be
qualified (per-path guardrails in the supplement).

\subsection{Splits, Protocol, and Preregistered Gates}

Calibration and held-out data are separated by session and cell,
and repetitions between two device resets never cross the split: the recipe calibrates on the single- and multi-light paths at both rungs plus
the $R_{\mathrm{soft}}$ sensor path and the cross-device cells, and
holds out the sensor-to-effect path at $R_{\mathrm{phys}}$, the
pairing that carries the main prediction claim; unseen delays
and actions serve as lesser feasibility levels, and an
unseen simulator adapter is instantiated later by the append-only
extension without reopening the physical program.  The main campaign
spans nine valid calibration sessions, 586 trials, and 1{,}070
eligible source-cleared pairs.  The protocol starts at 10 repetitions
per condition, finalizes counts from pre-freeze block variance, and
requires at least 3 qualified physical sessions on the principal
path.  Session-aware paired bootstraps summarize $\Delta$Brier with
win/tie/loss, and exact intervals keep every path and session count
visible.  Adaptive test selection stays outside the frozen campaign,
so it cannot contaminate the held-out gate.

\subsection{Frozen Predictors}

Table~\ref{tab:baselines} lists the 6 frozen predictors as a ladder
of structural granularity: from trusting the source's verdict
outright, through property-blind rates and a per-device lookup, to
the full property $\times$ stage/direction $\times$ path profile.

\begin{table}[t]
  \centering
  \caption{Frozen predictors. Cohort~1's preregistered gate required
  lower held-out Brier than strong lookup and path-only; cohort~2
  registered path-only as primary.}
  \label{tab:baselines}
  \footnotesize
  \begin{tabular}{cll}
    \toprule
    \# & Predictor & Calibration information used \\
    \midrule
    1 & Simulator verdict & constant risk 0.01 per sim pass \\
    2 & Global FCR & one Jeffreys rate per property \\
    3 & Scalar latency & pooled effect latency by boundary \\
    4 & Strong lookup & quantiles per action/direction/device \\
    5 & Path-only & rates per path and boundary \\
    6 & SimVerity & property $\times$ stage/dir.\ $\times$ path \\
    \bottomrule
  \end{tabular}
\end{table}

\subsection{Agentic Replication and Second Simulator}

Frozen plans isolate verdict transfer; live agents test whether the
same evaluation contract remains applicable when actions are selected
dynamically.  The audited object is an agent's deployment clearance,
not a policy learned by SimVerity.  The study uses one local
ReAct-style tool agent and one architecturally distinct
planner--executor agent, frozen at temperature zero, with verdict
metrics computed both conditional on matched actions and over the
natural live-agent distribution; this is not a model leaderboard: the
pre-registered auditability gate asks whether at least 80\% of traces
stay eligible and property-conditioned validity stays measurable
under dynamic tool selection.  After physical freeze, an append-only
registration qualified a second public simulator, S5-HES
\citep{siriweera2026s5hes}, and paired both simulators' frozen-grid
verdicts retrospectively against frozen P1 anchors (stored physical
outcomes); an orchestration study compares custom ReAct, an
unmodified production harness \citep{nousresearch2026hermes}, and the
planner under one local configuration, and a frontier contrast
repeats custom ReAct under another.  Table~\ref{tab:configs} binds
each measured or blocked cell to its exact model, route, and frozen
count; the frontier intervention changes weights, provider endpoint,
and serving stack together, so its unit is the full
model--client/serving configuration.  Stored pairs also compare
single-simulator, consensus, and either-pass policies; physical
evidence and the one-shot evaluator were never rerun.

\begin{table}[t]
  \centering
  \caption{Executable agent configurations; rows index
  model--client/serving configurations, not model-only comparisons.}
  \label{tab:configs}
  \scriptsize
  \setlength{\tabcolsep}{2.25pt}
  \renewcommand{\arraystretch}{0.97}
  \begin{tabularx}{\columnwidth}{@{}
    >{\hsize=.79\hsize\raggedright\arraybackslash}X
    >{\hsize=1.47\hsize\raggedright\arraybackslash}X
    >{\hsize=.74\hsize\raggedright\arraybackslash}X
    c l@{}}
    \toprule
    Config. & Model & Route / client & $n$ & State \\
    \midrule
    \multicolumn{5}{@{}l}{\textbf{Physical deployment}} \\
    M2 ReAct & gpt-oss:20b & Ollama; native & 32$\times$2 & invalid \\
    M3 planner & deepseek-r1:32b & Ollama; native & 32 & valid \\
    \cmidrule(lr){1-5}
    \multicolumn{5}{@{}l}{\textbf{Simulation robustness (frozen 32-cell grid)}} \\
    ReAct & gpt-oss:20b & local \texttt{/v1}; custom & 128 & measured \\
    Hermes v0.14.0 & gpt-oss:20b & local \texttt{/v1}; Hermes & 128 & measured \\
    Planner & gpt-oss:20b & local \texttt{/v1}; custom & 128 & measured \\
    ReAct & grok-4.5 & xAI API; custom & 128 & measured \\
    ReAct / Hermes & gpt-5.6-sol & OpenAI API & 0 & blocked$^{a}$ \\
    Hermes v0.14.0 & grok-4.5 & xAI API; Hermes & 0 & blocked$^{b}$ \\
    \bottomrule
  \end{tabularx}
  \begin{minipage}{\columnwidth}
    \vspace{2pt}
    \scriptsize
    Simulation rows: two trial orders, 12-turn budget; physical rows:
    frozen 32-trial manifests; every executed row enforces
    temperature 0.  Blocked rows ($n=0$): $^{a}$provider rejects
    non-default temperature; $^{b}$Hermes's custom-provider path
    fails silently and its native path exposes no enforceable
    identity.
  \end{minipage}
\end{table}

%% file: sections/evidence_and_calibration.tex
\section{Evidence and Calibration}
\label{sec:evidence}

\begin{figure*}[t]
  \centering
  \includegraphics[width=0.7\textwidth]{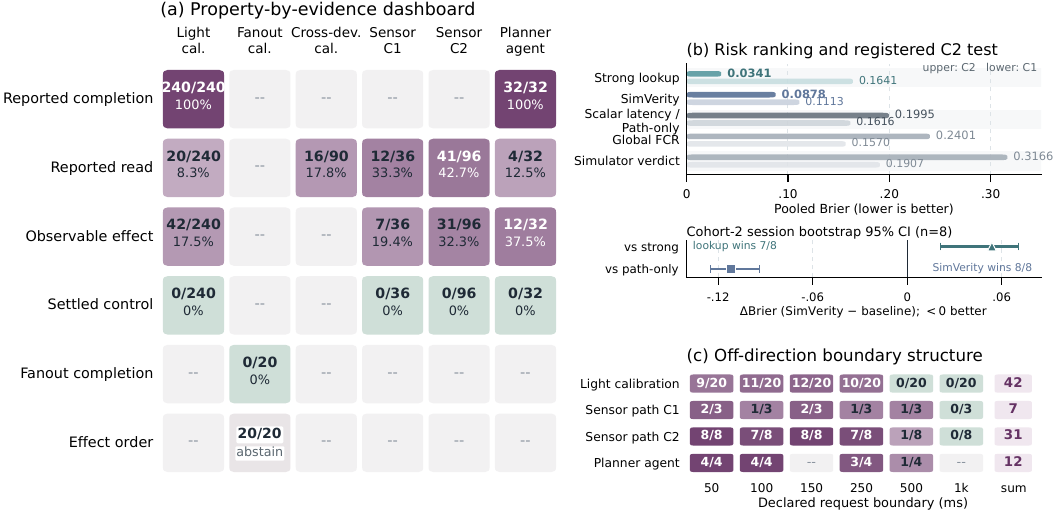}
  \caption{False clearances are property and boundary-selective.
(a) Cards report false clearances/source-cleared eligible pairs.
(b) Frozen Brier scores, cohort~2 upper and cohort~1 lower; scalar
latency and path-only coincide (registered comparisons in
Table~\ref{tab:prediction}). (c) Off-direction counts by boundary,
with row sums at right.}
  \label{fig:main-results}
\end{figure*}

\subsection{Calibration: Pilot and Main Campaign}
\label{sec:campaign}

A precommitted pilot paired SimuHome, a single smart light, and a
session-qualified camera witness (40 calibration and 30 held-out
trials, 0 abstentions, predictor frozen first, all 22 registered
checks independently reproduced): completion failed every trial, the
settled control none, and the five held-out observable false
clearances, all \texttt{off} trials at the unseen 0.1-s boundary, are
the introduction's counterexample.  The frozen profile's held-out
Brier ran an order of magnitude below every pilot baseline: a
feasibility result only, not satisfying the strong-baseline gate. The main campaign executed nine valid calibration sessions under the
registered manifest: four at $R_{\mathrm{soft}}$ and five at
$R_{\mathrm{phys}}$, totaling 586 trials and 1{,}070 eligible
simulator-pass pairs; exploratory pre-freeze sessions are excluded from every
estimate.  

\paragraph{Property-selective calibration structure.}
Across two devices and a grid extending to 0.05~s, the same executions yield
four different conclusions (Figure~\ref{fig:main-results}(a,c) and
Table~\ref{tab:transfer-ledger}).  Settled effect is uniformly valid;
observable failures are \texttt{off}-only through 0.25~s; reported state has
its own sub-50~ms boundary; and completion is uniformly invalid at its declared
read. 

\paragraph{Ambient light decides who can testify.}
The motivating platform (Google Home) watches homes through cameras
and a home's light changes all day, so physical sessions spanned the
deployment's real lighting day, morning to dark.  Ambient light then
exercised the witness checks in both directions: two early calibration
sessions were invalidated by their own pre-registered checks (optical
separation between targets failed at dusk, on--off contrast at midday),
and later agentic days refused or invalidated midday wash, dusk drift,
full-dark infrared, and sunlit saturation.  By the confirmatory cohort,
ambient light had been promoted from nuisance to a curtain-controlled
registered factor (next section).

\paragraph{Pre-registered replication blocks.}
Under a registered replication manifest, a television's reported state stayed stale after external power-on in every window (camera-based lower bounds near 160~s in the two cleanly anchored windows; supplement).  Cold \texttt{turn\_on} failed and warm retries succeeded intermittently; a success also repaired the
stale reported state: acting is what made reading correct.  A
registered probe confirmed event multiplication (85 recorded
transitions from 20 physical door-contact cycles); two suspected
mechanisms, event merging and detector dead-time, were tested and ruled
out.

\begin{table*}[t]
  \caption{Verdict-transfer ledger: false clearances/source-cleared
eligible pairs; ``--'' = not evaluated. Cross-simulator rows reuse
frozen P1 anchors. $\dagger$At 0.05~s, S5-HES leaves 120/160 reported
and observable replay pairs.}
  \label{tab:transfer-ledger}
  \centering
  \scriptsize
  \setlength{\tabcolsep}{4.0pt}%
  \renewcommand{\arraystretch}{0.97}
  \begin{tabular}{lllccccc}
    \toprule
    Block & Verdict source & Deployment pairing & Audit object & Completion & Reported & Observable & Settled \\
    \midrule
    P1 calibration & SimuHome & live $R_{\mathrm{phys}}$ & plan (240) & 240/240 & 20/240 & 42/240 & 0/240 \\
    P5c calibration & SimuHome & live $R_{\mathrm{phys}}$ & plan (90) & -- & 16/90 & -- & -- \\
    P3 held-out (one-shot) & $R_{\mathrm{soft}}$ & held-out $R_{\mathrm{phys}}$ & plan (36) & -- & 12/36 & 7/36 & 0/36 \\
    M3 live agent & SimuHome & live $R_{\mathrm{phys}}$ & M3 (32/32 match) & 32/32 & 4/32 & 12/32 & 0/32 \\
    Cross-sim replay & SimuHome & frozen P1 anchors & plan (160) & 160/160 & 20/160 & 30/160 & 0/160 \\
    Cross-sim replay & S5-HES & frozen P1 anchors & plan (160) & 160/160 & 0/120$^\dagger$ & 21/120$^\dagger$ & 0/160 \\
    \bottomrule
  \end{tabular}
\end{table*}

\subsection{Does the Audit Port?  Site Replications and Scope}

To test whether the audit and its discipline install beyond their
birthplace, two further deployments, independently instrumented and
preregistered on different device stacks, repeated the binary
relay--lamp audit: Site B, a separately operated office with variable
lighting conditions and a webcam (five qualified sessions, 120
trials), and Site C, a second home on a different vendor stack with an
Apple TV, bedroom lamps, and a home security camera (two software-only
sessions, 80 trials).  Nothing is pooled across sites and the frozen
predictor was never invoked: these sites are the paper's cross-stack
evidence that the audit itself ships, its protocol and discipline
running unchanged on deployments we did not build.  Every completion,
reported-state, and settled-postcondition pair passed: zero false
clearances in their trials.  The zeros are mechanism, not luck, and
the mechanism is one simulators do not model: how an integration
reports state.  Both sites' relay integrations are confirmation-gated
(reported state commits only on device acknowledgment), leaving no
window to false-clear; Site C also measured the opposite, optimistic
reporting that runs 50--500 ms ahead of the device.  Instead of sites,
semantics decides where reported state can lie.  Cameras at both sites could not certify
sub-second timing, so short-boundary optical cells abstained, exactly
as the contract requires (per-site setups and trace bridges in the
supplement).  What the sites do not show is equally simple: they say
nothing about how often such failures occur elsewhere, and they never
ran the risk profile or the agents; the prediction evidence remains
one home and one held-out pairing, 132 trials over eleven sessions.
Everything beyond this operational scope rests on the
one-shot, agentic, and cross-simulator tests reported next.

%% file: sections/held_out_evaluation.tex
\section{Results}
\label{sec:results}

\subsection{Property-Selective Fidelity}

The calibration card contains 1{,}090 rows: 1{,}070 binary pairs and
20 effect-order abstentions (camera cadence cannot resolve ordering).
Within the same executions, completion can fail universally while
settled effect never does, and both reported and observable verdicts
vary by path.  A scalar ``task success'' score would erase both the
property failure and the coverage boundary.

\subsection{Predicting False Clearance}

After freezing six predictors, the held-out path--rung pairing ran once
as an initial cohort: 36 source-cleared trials across three qualified
$R_{\mathrm{phys}}$ sensor-to-effect sessions (a door contact
dispatching a lamp).  Verdicts on this path originate at
$R_{\mathrm{soft}}$, not the simulator, so the held-out test asks
whether software-rung clearances survive the physical edge.  Failures
remained selective by direction and observation channel, not a uniform
path defect (Table~\ref{tab:transfer-ledger});
Figure~\ref{fig:main-results}(b) and Table~\ref{tab:prediction} carry
the registered comparisons, and both prediction arrays are released
for recomputation. All predictors covered 36/36; because the path--rung pairing was held
out, path-only and SimVerity both predicted through frozen level-1
backoff, so the comparison tests whether the declared
structure (Table~\ref{tab:baselines}) survives backoff.
The profile won all three initial sessions against path-only and its
pre-registered point-Brier gate passed; against strong lookup it was
point-better with resampling support spanning zero (supplement).  A
sign test cannot resolve three sessions ($p{=}0.125$), the thinness
the confirmatory cohort was registered to close.

A preregistered confirmatory second cohort reran the same pairing:
eight curtain-controlled sessions in one day (three ambient regimes),
the same frozen predictor, the first cohort sealed, one evaluation written
into a separate artifact.  All eight attempts qualified, and the 96 fresh trials
preserved the structure (observable FCR 31/96, reported 41/96, settled
0/96, 0 abstentions).  The registered primary criterion was met
decisively: 8/8 session wins against path-only, sign and exact
Wilcoxon $p{=}0.0039$; across both cohorts the profile wins all eleven
held-out sessions.  The registered secondary comparison reversed:
strong lookup beat the profile in seven of eight sessions.  The two
cohorts bracket a condition-dependent comparison: under natural
ambient variation the structured profile was point-better with
undecided support; under stationary controlled regimes the per-cell
lookup memorized best.  As a hypothesis, memorization pays under
stationarity while declared structure is aimed at surviving condition shift.  Six of eight sessions produced identical
twelve-trial outcome vectors, reported openly; because the second
protocol was registered with the first cohort's results known, the
11-session tally is descriptive, not a prospective joint test.

\subsection{Auditability Under Live Agents}

\begin{figure*}[t!]
  \centering
  \includegraphics[width=0.8\textwidth]{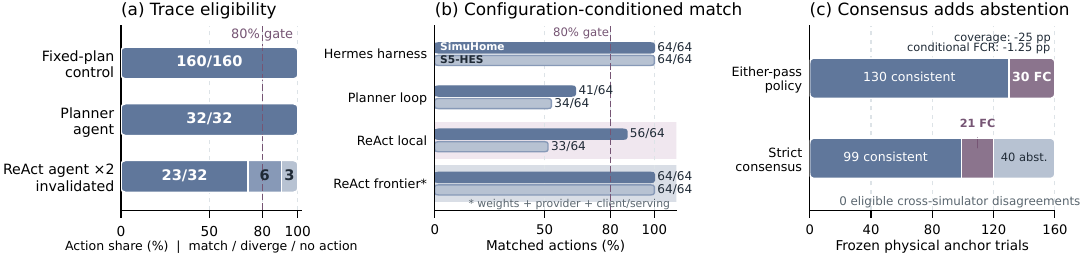}
  \caption{Auditability is configuration-conditioned; consensus changes
coverage instead of detection. (a) Fixed-plan and planner traces fully
match; both ReAct runs share a 23/6/3 matched/divergent/no-action
split (no M2 rate). (b) Match rates pool two sessions per executable
configuration, not model weights alone. (c) Zero eligible
disagreements on 160 anchors: strict consensus lowers conditional FCR
by 1.25 points by surrendering 25 coverage points.}
  \label{fig:agentic-boundaries}
\end{figure*}

On frozen 32-trial manifests, the live-agent study replaces the plan with
the M2 (ReAct) and M3 (planner--executor) physical configurations indexed in
Table~\ref{tab:configs}.  Capture arms at the agent's final
state-changing action; the witness capture time lands 0.393--0.739~s
after the requested boundary in M3, neither the declared delay nor the
capture time is hidden physical-effect time, and bounded capture under
monotone dimming can only undercount short-boundary false clearances
(supplement).
Figures~\ref{fig:main-results}(c) and~\ref{fig:agentic-boundaries}(a) report
the boundary cells and auditability outcome; dry runs and gate-invalidated
sessions are excluded from every estimate.

\begin{table}[t]
  \centering
  \caption{Held-out prediction tests.  $\Delta$ = SimVerity minus
  baseline (negative favors SimVerity); cohort-1 intervals are
  resampling support, cohort-2 bootstrap 95\% CIs.  Cohort 2
  preregistered path-only as primary, strong lookup as secondary (no
  threshold).}
  \label{tab:prediction}
  \scriptsize
  \setlength{\tabcolsep}{1.3pt}
  \renewcommand{\arraystretch}{0.97}
  \begin{tabular}{@{}lcc@{}}
    \toprule
     & Cohort 1 ($n{=}3$) & Cohort 2 ($n{=}8$) \\
    \midrule
    Simulator-verdict Brier & 0.1907 & 0.3166 \\
    Path-only Brier & 0.1616 & 0.1995 \\
    Strong-lookup Brier & 0.1641 & 0.0341 \\
    SimVerity Brier & 0.1113 & 0.0878 \\
    \addlinespace[1pt]
    Wins vs path-only & 3/3 ($p{=}0.125$) & 8/8 ($p{=}0.0039$) \\
    $\Delta$ vs path-only & $[-0.0940,-0.0077]$ & $[-0.1251,-0.0936]$ \\
    Wins vs strong lookup & -- & 1/8 \\
    $\Delta$ vs strong lookup & $[-0.1724,+0.0203]$ & $[+0.0212,+0.0707]$ \\
    \bottomrule
  \end{tabular}
\end{table}

\paragraph{The pattern survives an architecturally distinct live agent.}
\looseness=-1
M3 passed all gates with zero abstention.  Its observable failures remain
\texttt{off}-only and boundary-graded, while reported state reproduces the
sub-50~ms boundary (Figure~\ref{fig:main-results}(a,c)).  The auditability gate therefore passes: the pattern remains measurable
under dynamic tool selection; rates are not compared with the frozen-plan
anchor because boundaries reference the agent's own final action.  The
planner's immediate return reads pre-update state (32/32 completion
failures), whereas ReAct's closing turn lands after the update.

\paragraph{Auditability is itself configuration-conditioned.}
\looseness=-1
M2 failed twice with identical failed-trial sets;
Figure~\ref{fig:agentic-boundaries}(a) shows the registered
decomposition.  With the shared local-model configuration fixed
(Table~\ref{tab:configs}), Figure~\ref{fig:agentic-boundaries}(b)
separates architectural labels from executable configurations: the
production harness remains fully matched across both simulators while
the custom loops cross the auditability gate in a simulator-dependent
way (loss classes in the supplement).  Switching the fixed custom ReAct loop and task grid
to the registered frontier configuration restores every action match.  This
pre-registered full-configuration intervention falsifies ReAct itself as the
stable cause, but cannot separate model weights from provider, endpoint, and
serving behavior.  Auditability is therefore a property of the executable
configuration, not of a harness label.

\subsection{Cross-Simulator Transfer and Clearance Policies}

\looseness=-1
After physical freeze, a second simulator, S5-HES, qualified through its
engine below the built-in LLM \citep{siriweera2026s5hes}; its 100-ms tick
structurally abstains at 0.05~s.  The simulators never disagreed when mutually eligible,
and Table~\ref{tab:transfer-ledger} shows both inherit the same
property-selective physical failures.  Figure~\ref{fig:agentic-boundaries}(c)
exposes the policy result: strict consensus equals the stricter
simulator, either-pass the permissive one; the apparent risk reduction
is entirely forced abstention at the coarser clock, not detection by a
second judge.  On this grid the second green light supplied no
independent opinion, and deployment anchors alone calibrate risk; bridged
onto two further preregistered deployments,
the same frozen clearances incur zero reported or settled false
clearances (supplement).  Across
these studies, no simulator-side substitute suffices: simulator
agreement and stronger executable configurations each fail to replace
deployment-grounded verdict assurance.

%% file: sections/related_work.tex
\section{Related Work}
\label{sec:related}

\paragraph{Simulation validity and verdict preservation.}
Fit-for-purpose validity \citep{sargent2013verification}, CPS conformance
\citep{abbas2014formal,roehm2019conformance}, and digital-twin trace alignment \citep{munoz2024measuring} formalize model trust; the L0--L4 ladder names verdict transfer as its highest rung \citep{oefinger2026validate}, which SimVerity instantiates as per-property agreement, FCR, abstention, and held-out prediction.

\paragraph{Sim--real predictivity and physical testing.}
Sim-to-real studies test whether rankings predict reality \citep{kadian2020predictivity,li2024simpler,truong2022rethinking}, world-model evaluators automate rollout scores \citep{quevedo2025evaluating,li2025worldeval}, and driving studies compare virtual and physical failures \citep{fremont2020formal,stocco2022mind,haq2021offline}; the per-scenario test-track comparison is the closest verdict-level precedent, and SimVerity makes VF/FCR reusable with probability quality tested.

\paragraph{Agent benchmarks and sandbox validity.}
Smart-home and tool agents receive saved-state or sandbox verdicts
\citep{rivkin2024sage,li2025homebench,seo2026simuhome,li2026smhbench,
gu2026homeflow,yao2024taubench}, yet offline evaluation can mislead
deployment \citep{kapoor2024agents,xue2025illusion}.  ToolEmu validates sim-fail to
plausible-real-fail \citep{ruan2024toolemu}; FCR asks whether sim-pass becomes
observed deployment-fail.  LLM personas and virtual ambient sensors synthesize smart-home traces \citep{juttner2026simulating,leng2026agentsense}, enriching simulators whose verdicts still need auditing; a sim-to-real agenda and failure-injection benchmark \citep{liu2026gap,zhou2026simulationlies} motivate SimVerity's paired measurement layer.
\paragraph{Evaluation validity as an alignment problem.}
Alignment-track work audits the evaluation pipeline itself:
contamination, steerability side effects, evaluator bias, and
success--safety separation
\citep{liang2026arxivroll,chang2026steerability,recchia2026confirmationbias,lee2026mobilesafetybench}.
BenchGuard audits benchmark artifacts and DFAH measures replay
determinism \citep{tu2026benchguard,khatchadourian2026replayable};
neither tests whether a simulator verdict survives witnessed
deployment.  Deployment simulation forecasts prevalence from prior
traffic \citep{williams2026predicting} and Composable Assurance
propagates formal assertions through MLOps \citep{zhao2026composable};
SimVerity instead calibrates whether simulator verdicts remain
deployment-admissible.

%% file: sections/limitations_ethics_reproducibility.tex
\section{Limitations, Ethics, and Reproducibility}
\label{sec:limitations}

The primary predictor profile covers one held-out physical path--rung
pairing, eleven sessions in two cohorts: it estimates risk, not
certification, and the strong-lookup
comparison splits by ambient condition, favoring the lookup under
stationary control.  The unpooled site cohorts test portability alone.
Physical evidence has one valid planner configuration; invalid ReAct sessions
mark a configuration-level auditability boundary. The model--client/serving intervention cannot isolate weights from serving, with frontier cells blocked (Table~\ref{tab:configs}); ecosystems without first-class determinism resist frozen evaluation. Only whitelisted lights and isolated proxies were tested.  Frames are reduced to scalars and discarded; no human data are collected. Versioned manifests, sanitized ledgers, hashes, and replay code regenerate all aggregates (abstentions/invalid sessions retained);
frozen arrays recompute one-shot metrics; materials provided.

%% file: sections/conclusion.tex
\section{Conclusion}
\label{sec:conclusion}

A green check from simulation is a promise about the physical world; we measured how often it holds. On a live home, the same passes split into four verdicts, and a camera kept catching lights that had not yet obeyed. The blind spots proved predictable: a risk profile frozen before testing anticipated them on a path it never physically measured, beating a property-blind baseline in all eleven held-out sessions across two cohorts. Auditability followed the executable configuration, and a second qualified simulator never disagreed with the first: only physical measurement exposed their shared blind spots. Ship the audit, not the answer: answers are deployment-local by construction.